\documentclass[10pt, logo, onecolumn]{shengshu}

\usepackage[utf8]{inputenc}
\usepackage[T1]{fontenc}
\usepackage{charter}

\newcommand{\bffont}{\fontsize{8.8}{10}\selectfont}
\DeclareTextFontCommand{\textbf}{\bffont\bfseries\selectfont}

\usepackage{wrapfig}
\usepackage{xspace}
\usepackage{textgreek}
\usepackage{bm}
\usepackage{cleveref}
\usepackage{multirow}
\usepackage{subcaption}
\usepackage{orcidlink}
\usepackage{footmisc}
\usepackage{algorithm}
\usepackage{algpseudocode}
\usepackage{listings}
\usepackage{csquotes}
\usepackage{marvosym}
\usepackage{nicefrac}
\usepackage{multicol}
\usepackage{tabto}
\usepackage{adjustbox}
\usepackage{dblfloatfix}
\usepackage{verbatim}
\usepackage{mathtools}
\usepackage{array}
\usepackage{bbm}
\usepackage{makecell}
\usepackage{siunitx}
\usepackage{pdflscape}
\usepackage{arydshln}
\usepackage{hhline}
\usepackage{diagbox}
\usepackage[square,sort,comma,numbers]{natbib}
\usepackage{fp}

\definecolor{linkc}{rgb}{0, 0.44, 0.74}
\definecolor{eqc}{rgb}{1, 0, 0}
\definecolor{newcitecolor}{rgb}{0,0.6,0}
\definecolor{mygreen}{RGB}{34,139,34}
\definecolor{mylightblue}{RGB}{0,162,230}
\definecolor{deepyellow}{RGB}{255, 215, 0}
\definecolor{catgray}{gray}{0.92}
\definecolor{pearDark}{RGB}{171, 195, 87}
\definecolor{codebg}{RGB}{245, 245, 245}
\definecolor{keywordcolor}{RGB}{0, 0, 153}
\definecolor{commentcolor}{RGB}{34, 139, 34}
\definecolor{stringcolor}{RGB}{163, 21, 21}
\definecolor{numbercolor}{RGB}{128, 128, 128}

\newcommand{\ours}{Causal Forcing++\xspace}

\def\onedot{\futurelet\@let@token\@onedot}
\def\@onedot{\ifx\@let@token.\else.\null\fi\xspace}

\makeatletter
\def\blfootnote#1{\xdef\@thefnmark{}\@footnotetext{\scriptsize #1}}
\makeatother

\hypersetup{
    colorlinks=true,
    urlcolor=shengshublue,
}

\let\cite\citep

\usepackage{amsmath,amsfonts,bm}

\def\eqref#1{equation~\ref{#1}}

\def\1{\bm{1}}

\def\vzero{{\bm{0}}}

\def\vepsilon{{\bm{\epsilon}}}

\def\vf{{\bm{f}}}

\def\vv{{\bm{v}}}

\def\vx{{\bm{x}}}

\def\mI{{\bm{I}}}

\DeclareMathAlphabet{\mathsfit}{\encodingdefault}{\sfdefault}{m}{sl}
\SetMathAlphabet{\mathsfit}{bold}{\encodingdefault}{\sfdefault}{bx}{n}

\def\gN{{\mathcal{N}}}
\def\gO{{\mathcal{O}}}

\newcommand{\E}{\mathbb{E}}

\newcommand{\KL}{D_{\mathrm{KL}}}

\title{Causal Forcing++: Scalable Few-Step Autoregressive Diffusion Distillation for Real-Time Interactive Video Generation}

\author{%
  \textbf{Min Zhao}$^{1,2~*}$, \textbf{Hongzhou Zhu}$^{1,2~*}$, \textbf{Kaiwen Zheng}$^{1}$, \textbf{Zihan Zhou}$^{3}$,
  \textbf{Bokai Yan}$^{3}$, \textbf{Xinyuan Li}$^{1}$, \textbf{Xiao Yang}$^{1}$, 
  \textbf{Chongxuan Li}$^{3}$, \textbf{Jun Zhu}$^{1,2~\dagger}$\\
  $^1$Tsinghua University\,
  $^2$ShengShu\,
  $^3$Renmin University of China\\
  {\small $^*$Equal contribution. \quad $^\dagger$Correspondence to: Jun Zhu.}\\
\texttt{\{gracezhao1997, suinibian74\}@gmail.com;}
\texttt{dcszj@tsinghua.edu.cn}
}

\begin{abstract}
Real-time interactive video generation requires low-latency, streaming, and controllable rollout. Existing autoregressive (AR) diffusion distillation methods have achieved strong results in the chunk-wise 4-step regime by distilling bidirectional base models into few-step AR students, but they remain limited by coarse response granularity and non-negligible sampling latency. In this paper, we study a more aggressive setting: frame-wise autoregression with only 1--2 sampling steps. In this regime, we identify the initialization of a few-step AR student as the key bottleneck: existing strategies are either target-misaligned, incapable of few-step generation, or too costly to scale. We propose \textbf{Causal Forcing++}, a principled and scalable pipeline that uses \emph{causal consistency distillation} (causal CD) for few-step AR initialization. The core idea is that causal CD learns the same AR-conditional flow map as causal ODE distillation, but obtains supervision from a single online teacher ODE step between adjacent timesteps, avoiding the need to precompute and store full PF-ODE trajectories. This makes the initialization both more efficient and easier to optimize. The resulting pipeline, \ours, surpasses the SOTA 4-step chunk-wise Causal Forcing under the \textit{\textbf{frame-wise 2-step setting}} by 0.1 in VBench Total, 0.3 in VBench Quality, and 0.335 in VisionReward, while reducing first-frame latency by 50\% and Stage 2 training cost by $\sim$$4\times$. We further extend the pipeline to action-conditioned world model generation in the spirit of Genie3.
Project Page: \textbf{\href{https://github.com/thu-ml/Causal-Forcing}{\textcolor{shengshublue}{https://github.com/thu-ml/Causal-Forcing}}} \& \textbf{\href{https://github.com/shengshu-ai/minWM}{\textcolor{shengshublue}{https://github.com/shengshu-ai/minWM}}}.

\end{abstract}

\begin{document}
\maketitle

\section{Introduction}
Video generation models are rapidly evolving from passive content generators into interactive world models~\cite{videoworldsimulators2024, bao2024vidu,wan2025wan,kong2024hunyuanvideo,yang2024cogvideox,lin2024open,zheng2024open,sun2025worldplay,genie3,huang2025live,ki2026avatar,sun2025streamavatar,feng2025vidarc,ye2026worldactionmodelszeroshot}, where low latency, streaming rollout, and user-controllable interaction are essential. Autoregressive (AR) diffusion models~\cite{jin2024pyramidal,teng2025magi,chen2025skyreels} are a natural fit for this goal, as they perform causal rollout across frames or chunks while retaining diffusion-based generation within each segment. Recent AR diffusion distillation methods~\cite{yin2025slow,huang2025self,zhu2026causal,huang2025live,sun2025worldplay} have achieved promising results by distilling bidirectional video diffusion models, such as Wan~\cite{wan2025wan} and Hunyuan~\cite{kong2024hunyuanvideo}, into few-step AR students. However, these methods typically rely on chunk-wise autoregression with 4-step sampling, which still falls short of real-time interaction due to coarse response granularity and non-negligible sampling latency.  We therefore push AR diffusion distillation to a more aggressive and largely underexplored regime: \emph{frame-wise autoregression with only 1--2 sampling steps}.

We identify the initialization of a few-step AR student before asymmetric DMD as the key bottleneck in this aggressive regime, where existing strategies fail in complementary ways. ODE initialization with a bidirectional teacher, as used in CausVid~\cite{yin2025slow} and Self Forcing~\cite{huang2025self}, is architecturally misaligned with causal rollout: the teacher trajectory depends on future frames that are unavailable to an AR student, thereby providing an incorrect regression target. Directly using a multi-step AR diffusion model for initialization, as in LiveAvatar~\cite{huang2025live} and WorldPlay~\cite{sun2025worldplay}, avoids this mismatch but lacks few-step generation capability; under frame-wise 1--2 step generation, its per-frame approximation error is severely amplified during self-rollout. Causal ODE initialization, as in Causal Forcing~\cite{zhu2026causal}, corrects the learning target by distilling from an AR teacher, but requires generating full multi-step PF-ODE trajectories for every training sample, making it costly to scale. Therefore, a satisfactory initialization for this regime must be simultaneously \emph{AR}, \emph{few-step}, and \emph{scalable}.

To this end, we introduce \textbf{Causal Forcing++}, a principled and scalable pipeline that uses \emph{causal consistency distillation} (causal CD) for few-step AR student initialization. Our key observation is that causal ODE distillation and causal CD aim to learn the same object: the AR-conditional flow map (or namely the consistency function~\cite{song2023consistency,wang2024phased,lu2024simplifying,zheng2025large}) of the teacher. They differ, however, in how the supervision is obtained. Causal ODE distillation requires the AR teacher to generate an entire multi-step PF-ODE trajectory for each training sample, which must be precomputed and stored offline. In contrast, causal CD obtains supervision from a single online teacher ODE step between adjacent timesteps on real videos. Therefore, causal CD serves as a principled substitute for causal ODE initialization, while avoiding the expensive trajectory-generation bottleneck. Beyond efficiency, this local supervision also improves quality: adjacent-timestep consistency yields a smaller per-step optimization gap than causal ODE distillation, which regresses noisy intermediate states directly to clean endpoints~\cite{liu2023instaflow}. As a result, causal CD is easier to optimize and empirically produces a stronger few-step AR student.

Experiments on Wan2.1-1.3B~\cite{wan2025wan} validate the effectiveness of Causal Forcing++ in this aggressive low-latency regime. Under frame-wise 2-step generation, Causal Forcing++ achieves the best overall performance among existing AR diffusion distillation methods, improving VBench Total, VBench Quality, and VisionReward over prior methods while reducing first-frame latency by 50\%. Ablation studies further show that causal CD consistently matches or outperforms causal ODE initialization across 1-step, 2-step, and 4-step settings, while reducing the Stage 2 cost by about $4\times$ and requiring no auxiliary trajectory storage. We also examine causal score-distillation initialization and find that, although it can produce sharper early frames, its mode-seeking behavior makes it more sensitive to accumulated history errors during AR rollout, leading to stronger exposure bias. Finally, we demonstrate that Causal Forcing++ naturally extends to action-conditioned world model generation by distilling a camera-pose-conditioned generator into an interactive AR world model.

\begin{figure}[htbp]
    \centering
    \vspace{-.25cm}
    \includegraphics[width=\textwidth]{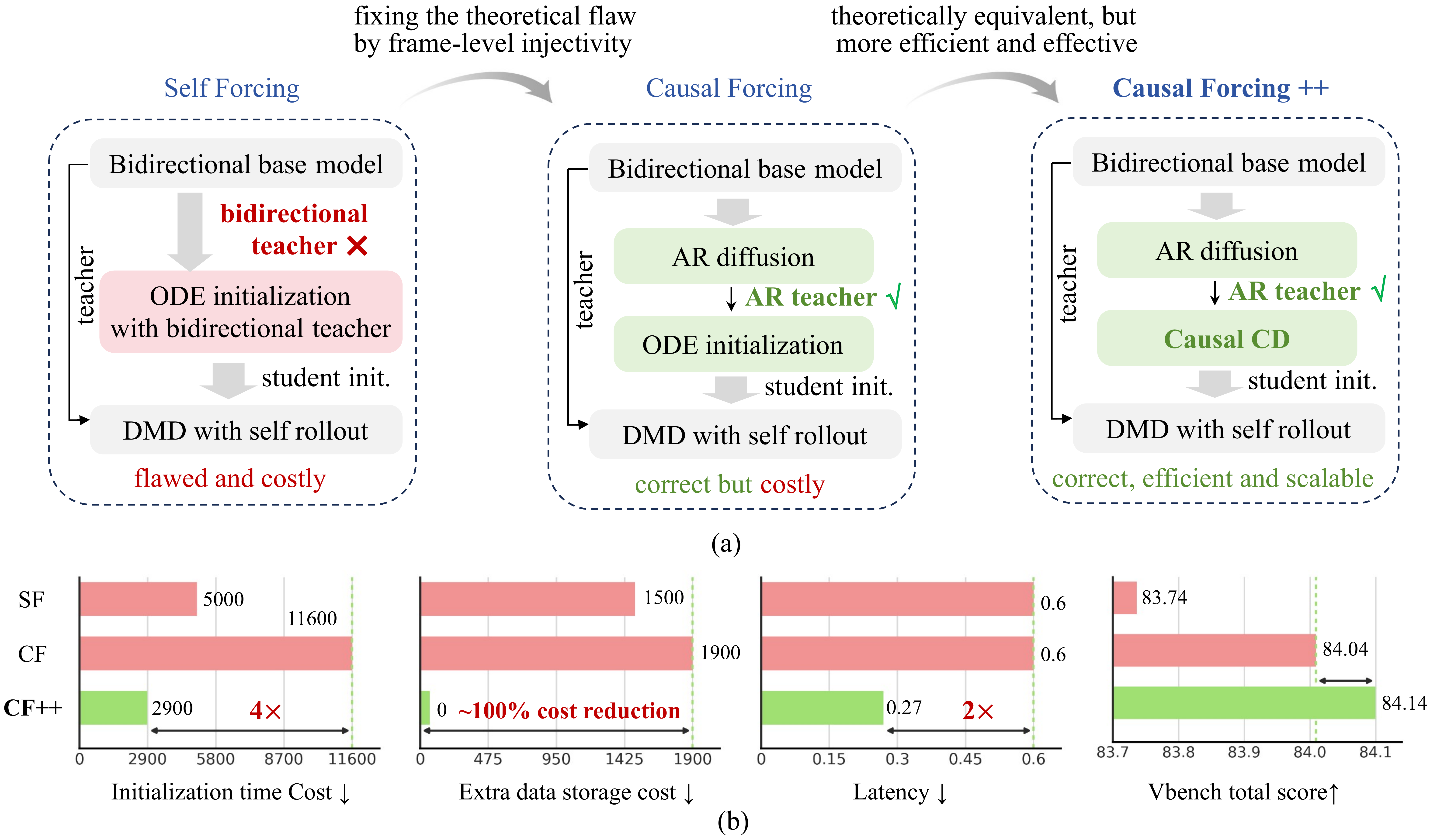}
    \vspace{-.2cm}
    \caption{\textbf{Overall framework of our Causal Forcing++ and the comparison with existing methods.} (a) Causal Forcing (CF) fixes Self Forcing (SF)'s frame-level injectivity flaw but remains costly; our Causal Forcing++ (CF++) is theoretically sound, efficient, and scalable. (b) Our CF++ reduces training cost by 4$\times$, requires no extra data curation, and achieves 50\% lower latency and higher VBench scores than SF and CF.}
    \label{fig:overall}
    \vspace{-.5cm}
\end{figure}

\section{Background}
\subsection{Autoregressive Diffusion for Interactive Video Generation}
\label{sec: diffusion}

\paragraph{Generative modeling through diffusion.}
Score-based diffusion models~\cite{ho2020denoising, song2020score} typically perturb the data $\vx_0\sim p_{\mathrm{data}}(\vx_0)$ through the diffusion process $\vx_t = \alpha(t)\vx_0 + \sigma(t)\vepsilon, \, \vepsilon \sim \gN(\vzero, \mI), t\in [0,1]$, where $\alpha(\cdot),\sigma(\cdot)$ are functions of the timestep $t$, namely the noise schedules. Trained in this process, the model learns score-related information about $p_{\mathrm{data}}(\vx_0)$. A recent trend is flow matching~\cite{liu2022flow, lipman2022flow}, which sets $\alpha(t)=1-t$ and $\sigma(t)=t$, and parameterizes the model using velocity prediction, denoted by $\vv_\theta$. The model is trained by minimizing $\E_{\vx_0,\vepsilon, t}[||\vv_\theta(\vx_t,t)-\vv_t||^2]$, where $\vv_t := \frac{\mathrm{d}\vx_t}{\mathrm{d}t} = \vepsilon - \vx_0$. By using the optimal model $\vv_\theta$ to solve the probability flow ordinary differential equation (PF-ODE)~\cite{song2020score} $\mathrm{d}\vx_t = \vv_\theta(\vx_t,t)\mathrm{d}t,\,t:1\rightarrow 0$, one can generate samples $\vx_0 \sim p_{\mathrm{data}}(\vx_0)$. Owing to their remarkable expressive power, diffusion models have achieved widespread success in image and video generation~\cite{zhao2022egsde,zhao2024identifying,bao2024vidu,zhao2025controlvideo,zhao2025riflex,zhao2025ultravico,zhao2025ultraimage}.

\paragraph{Autoregressive diffusion for video generation.}
Despite the remarkable success of diffusion models in video generation~\cite{videoworldsimulators2024,lin2024open,zheng2024open,yang2024cogvideox,wan2025wan,kong2024hunyuanvideo,bao2024vidu,li2025radial}, they are typically bidirectional, generating the entire video content in a single pass. This leads to high latency, namely a long delay from the start of generation to the completion of the first frame, and also makes them non-interactive, since the user must specify the full conditioning signals in advance. By contrast, autoregressive (AR) generation~\cite{wu2021godiva,hong2022cogvideo,wu2022nuwa,weissenborn2019scaling,yan2021videogpt,zhao2025ultraimage,zhao2025riflex,deng2024autoregressive,kondratyuk2023videopoet} operates on much smaller generation units, thereby offering lower latency. It also enables interactive generation, as users can provide feedback based on the content already generated and adjust subsequent conditioning signals accordingly. To combine the high generation quality of diffusion models with the low latency and interactivity of AR models, a recent trend is AR diffusion~\cite{jin2024pyramidal,teng2025magi,chen2025skyreels}, which performs autoregression across frames (or chunks) and diffusion within each frame (or chunk). These models typically adopt a causal attention mask to be trained with teacher forcing or its variants~\cite{teng2025magi, chen2024diffusion, wu2025pack, guo2025end, po2025bagger}, and perform self-rollout inference with a KV cache.

\subsection{Autoregressive Diffusion Distillation}
\label{sec:bg-distillation}
Although AR diffusion enables interactivity, its multi-step generation process still hinders real-time generation. This has motivated the development of AR diffusion distillation~\cite{lin2025autoregressive,lin2025diffusion,yang2025towards,lu2025reward,wang2026worldcompass,yang2025longlive,liu2025rolling,cui2025self,sun2025worldplay}.

\paragraph{CausVid.}
CausVid~\cite{yin2025slow}, one of the earliest representatives of the current AR diffusion distillation paradigm, adopts a two-stage framework:
\emph{\textbf{(1) ODE initialization}}, which samples PF-ODE trajectories from a bidirectional diffusion model and trains an AR student via regression;
\emph{\textbf{(2) asymmetric DMD}}~\cite{wang2023prolificdreamer,luo2023diff,yin2024one}, which keeps the teacher and critic bidirectional while using the ODE-initialized AR model as the student, trained under diffusion forcing on real data.
The rationale is that bidirectional diffusion models currently achieve stronger performance than AR diffusion models, so a stronger bidirectional teacher is expected to transfer a better generative distribution to the AR student.

\paragraph{Self Forcing.}
Self Forcing~\cite{huang2025self} improves upon CausVid by \textbf{correcting the asymmetric DMD stage}. Specifically, it observes that CausVid trains the student under diffusion forcing on real data, so each generated frame is conditioned on ground-truth context rather than on self-generated prefixes. Consequently, the resulting frames do not form a valid generated video when concatenated, leading to a substantial gap between training and inference-time self-rollout. To resolve this issue, Self Forcing replaces diffusion forcing generation in the DMD stage with student self-rollout, thereby aligning training with inference and substantially improving performance.

\paragraph{Causal Forcing.}
Causal Forcing~\cite{zhu2026causal} \textbf{corrects the ODE initialization stage} used in CausVid and Self Forcing, while retaining the asymmetric DMD design of Self Forcing. It points out that, unlike DMD which merely pursues distribution matching, ODE distillation is intended to match the generation trajectories. Consequently, an AR student is theoretically incapable of fitting the ODE trajectories induced by a bidirectional teacher, and thus cannot properly bridge the architectural gap. Motivated by this observation, Causal Forcing first fine-tunes a bidirectional diffusion model into an AR diffusion model, and then uses the resulting AR teacher to generate ODE trajectories for initializing the AR student. The initialized student is subsequently optimized with asymmetric DMD, while the teacher and critic remain bidirectional. 

In summary, the training pipeline consists of the following three stages, whose terminology we adopt below: \emph{\textbf{(1) Stage 1: multi-step AR diffusion training} via teacher forcing; \textbf{(2) Stage 2: causal ODE initialization} with the AR teacher; and \textbf{(3) Stage 3: asymmetric DMD} with student self-rollout.}

\section{Method}
\label{sec: method}

The AR diffusion distillation pipelines reviewed in Sec.~\ref{sec:bg-distillation} have achieved strong results in chunk-wise (typically 3 latent frames) 4-step AR generation, but two questions remain open. First, none of them has been validated in more aggressive low-latency regimes---in particular, \emph{frame-wise AR generation with as few as 1--2 sampling steps}---which would better realize the low-latency promise of real-time interactive generation. Second, the existing few-step initialization strategy, ODE distillation, requires the teacher to generate full PF-ODE trajectories for every training datum; this is structurally expensive and makes systematic exploration of harder regimes costly. In this section we address both. We first show that no existing initialization strategy is satisfactory in our target regime (Sec.~\ref{sec:necessity}); we then propose causal consistency distillation as a principled and scalable substitute for causal ODE distillation (Sec.~\ref{sec:substitute}); finally, we extend our method to action-conditioned world model generation (Sec.~\ref{sec:application}). The overview of our method and its relation to previous works are shown in Fig.~\ref{fig:overall}.

\subsection{The Necessity of Few-Step AR Student Initialization}
\label{sec:necessity}

\begin{figure}[t]
  \centering
    \includegraphics[width=\linewidth]{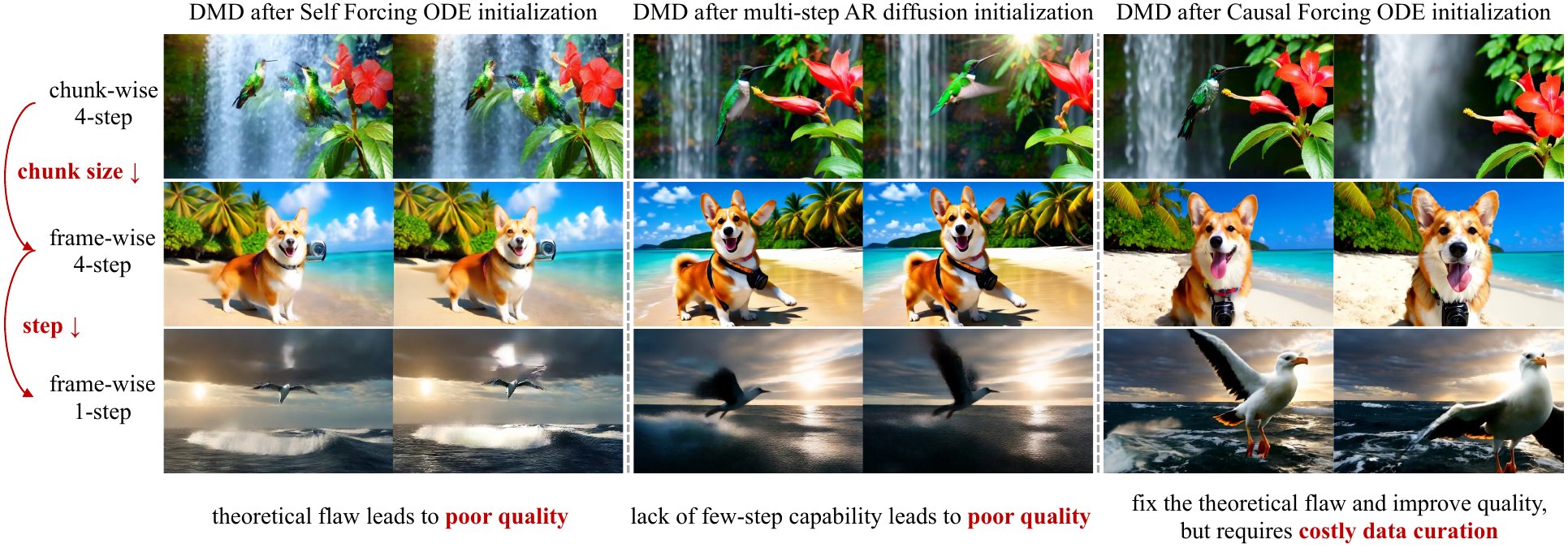}
  \caption{\textbf{Performance of existing initialization methods after DMD.} Self Forcing ODE initialization and multi-step AR diffusion initialization lead to poor quality and degrade as the setting becomes more aggressive, while Causal Forcing's causal ODE initialization performs well but is costly and therefore difficult to scale.}
  \label{fig:multi-step}
\end{figure}

Since asymmetric DMD is highly sensitive to its few-step student initialization~\cite{zhu2026causal}, we begin by examining whether existing initialization strategies suffice in our target regime. In recent works, three options have been proposed: (i) distilling a few-step AR student from a \emph{bidirectional} teacher via ODE distillation, as in CausVid~\cite{yin2025slow} and Self Forcing~\cite{huang2025self}; (ii) skipping few-step distillation entirely and using the multi-step AR diffusion model directly, as in LiveAvatar~\cite{huang2025live} and WorldPlay~\cite{sun2025worldplay}; and (iii) distilling a few-step AR student from an \emph{AR} teacher via ODE distillation, as in Causal Forcing~\cite{zhu2026causal}. We compare these three candidates as the autoregressive unit shrinks from chunk-wise to frame-wise and the sampling step count drops from 4 to 1. As shown in Fig.~\ref{fig:multi-step}, we find that all three fall short for complementary reasons: one is architecturally misaligned, one is too weak, and one is too costly to scale. This motivates an initialization that is simultaneously \emph{AR}, \emph{few-step}, and \emph{scalable}, as we now establish.

\newcommand{\leftfigwidth}{0.67\linewidth}
\newcommand{\rightfigwidth}{0.25\linewidth}
\newcommand{\figxshift}{-0.02\linewidth}
\newcommand{\figgap}{0.05\linewidth}

\paragraph{ODE initialization with a bidirectional teacher is misaligned.}

The first candidate uses \emph{ODE distillation with a bidirectional teacher}, which is the initialization mechanism behind CausVid~\cite{yin2025slow} and Self Forcing~\cite{huang2025self}. As Causal Forcing~\cite{zhu2026causal} shows, this choice violates the \emph{frame-level injectivity} required by an AR student---the same noisy frame can correspond to multiple clean frames under different future contexts---so ODE distillation no longer recovers the AR flow map but instead collapses toward the conditional expectation $\E[\vx_0^i \mid \vx_t^i, \vx_t^{<i}, t]$, producing a blurred and poorly aligned initialization. This mismatch becomes more damaging in our frame-wise low-step settings, where the student must repeatedly roll out from its own imperfect history. Quantitatively, asymmetric DMD with Self-Forcing-style initialization already collapses in the chunk-wise 4-step setting, becomes even worse in the frame-wise 4-step setting, and eventually catastrophically breaks down in the frame-wise 1-step setting, as illustrated in Fig. \ref{fig:multi-step}(column 1). \emph{ODE initialization with a bidirectional teacher} is therefore fundamentally flawed and not a viable foundation for low-latency AR diffusion distillation.

\paragraph{Multi-step AR diffusion initialization degrades sharply in aggressive settings.} Using the multi-step AR diffusion model directly as the student initialization yields consistently weaker results than explicit few-step distillation, and this gap widens as we move toward lower-latency generation. As shown in Fig.~\ref{fig:multi-step}(column 2 vs. column 3), the gap is moderate under chunk-wise generation, larger under frame-wise 4-step, and largest under frame-wise 1-step, where the model nearly collapses. The reason is structural: reducing the chunk size increases the number of AR calls needed to generate a video, while reducing the sampling step count increases the approximation error within each call. These two effects compound during self-rollout, amplifying exposure bias. In this sense, asymmetric DMD acts as a refiner rather than a complete trainer: if the initialization lacks few-step capability, DMD inherits an optimization burden it cannot reliably absorb.

\paragraph{Causal ODE initialization with an AR teacher works, but is difficult to scale.}
Causal Forcing~\cite{zhu2026causal} addresses the architectural mismatch of Self-Forcing-style ODE initialization by replacing the bidirectional teacher \emph{with an AR teacher} and performing causal ODE distillation between AR models. This produces a well-aligned few-step initialization and leads to strong chunk-wise 4-step results after asymmetric DMD, as illustrated in Fig. \ref{fig:multi-step}(column 3). Its limitation is not correctness but scalability: each training datum requires the AR teacher to generate a full multi-step PF-ODE trajectory (e.g., 48 steps per sample); these paired trajectories must be stored offline; and they must be regenerated whenever the teacher, data distribution, or chunk-size configuration changes. Worse, the cost grows with task difficulty: harder regimes typically demand more training data and longer trajectories, compounding the bottleneck.
At our 80K-video scale, this data curation along with the training costs roughly $11{,}600$ A800-GPU hours and $1{,}900$\,GiB of additional storage, as quantified in Tab. \ref{tab:ablation}. Thus, causal ODE initialization works in principle but imposes a structural scaling bottleneck that limits its practical reach.

Taken together, the three existing options leave us with no satisfactory initialization for asymmetric DMD in aggressive low-latency regimes: one is architecturally misaligned, one lacks few-step capability, and one is too costly to scale. We therefore need an initialization that is simultaneously \emph{AR}, \emph{few-step}, and \emph{scalable}. 

\subsection{Causal Forcing++: Causal CD as a Principled and Scalable Substitute for Causal ODE Initialization}
\label{sec:substitute}

Sec.~\ref{sec:necessity} established that the initialization of asymmetric DMD should be simultaneously \emph{AR}, \emph{few-step}, and \emph{scalable}, and that causal ODE distillation~\cite{zhu2026causal} satisfies only the first two. Our key observation is that causal ODE distillation and causal consistency distillation (CD) share the same learning target: the flow map (or the consistency function) of the AR teacher. Building on this equivalence, we propose causal CD as the few-step AR student initialization. We show that this substitution is principled---sharing the learning target of causal ODE distillation---and brings two practical advantages: it eliminates the offline-trajectory bottleneck and yields a stronger initialization via a smaller per-step optimization gap. We refer to the resulting pipeline as \textbf{\textit{Causal Forcing++}}.

\paragraph{Causal ODE distillation shares the same target as causal consistency distillation.}
Causal ODE distillation~\cite{zhu2026causal} collects intermediate states $\vx_t^i$ and corresponding clean outputs $\vx_0^i$ along the PF-ODE trajectory of the AR diffusion teacher, and trains the student by MSE regression with teacher forcing:
\begin{align}
    \theta^*=\arg\min_\theta \mathbb{E}_{\vx_{\mathrm{gt}}^{<i},\, t,\, i,\, \vx_t^i}\left[\,\|G_\theta(\vx_t^i,\vx_{\mathrm{gt}}^{<i}, t)-\vx_0^i\|^2\,\right].
\end{align}
The minimizer of this objective is the AR-conditional flow map of the teacher,
\begin{align}
    \vf_\phi:(\vx_t^i,\,\vx_{\mathrm{gt}}^{<i},\,t)\mapsto \vx_0^i,
\end{align}
which maps $\vx_t^i$ at any time $t$ to the PF-ODE sample $\vx_0^i$ of the teacher diffusion model $\phi$ at $t=0$. 
Recognizing $\vf_\phi$ as the AR-conditional analog of the consistency function~\cite{song2023consistency,song2023improved,wang2024phased,lu2024simplifying,zheng2025large}, we lift the standard CD objective to the AR setting via teacher forcing:
\begin{align}
\label{eq:causal-cd}
\theta^*=\arg\min_\theta \mathbb{E}_{\vx_\text{gt},\,\vepsilon,\,t,\,i}\Big[w(t)\,d\big(G_\theta(\vx_t^i,\vx_\text{gt}^{<i}, t),\; G_{\theta^-}(\hat{\vx}^i_{t-\Delta t},\vx_\text{gt}^{<i}, t-\Delta t)\big)\Big],
\end{align}
where $\vx_t^i$ is obtained from the ground-truth $\vx_\text{gt}^i$ via the forward diffusion process, $\hat{\vx}^i_{t-\Delta t}$ is obtained by a single ODE step from $\vx_t^i$ using the AR teacher conditioned on the ground-truth prefix $\vx_\text{gt}^{<i}$, $\theta^-$ is the EMA of $\theta$ with stop-gradient, $w(\cdot)$ is a timestep-dependent weight, and $d(\cdot,\cdot)$ is a distance under a pre-defined norm. Under the flow-matching parameterization of Sec.~\ref{sec: diffusion},
\begin{align}
G_\theta(\vx_t^i,\vx_\text{gt}^{<i},t) = \vx_t^i - t\,\vv_\theta(\vx_t^i,\vx_\text{gt}^{<i},t),
\end{align}
where $\vv_\theta$ is the neural network. Following standard CD analysis~\cite{song2023consistency}, the error between the optimal model $G_{\theta^*}$ and the target flow map $f_\phi$ (namely the consistency function) is bounded by the numerical error of the ODE solver and is therefore negligible:
\begin{align}
    \mathrm{sup}||f_\phi(\vx_t^i, \vx_{\mathrm{gt}}^{<i},t) - G_{\theta^*}(\vx_t^i, \vx_{\mathrm{gt}}^{<i},t)||_2 = \gO((\Delta t)^p),
\end{align}
where $\Delta t$ denotes the maximum difference between adjacent timesteps, and the ODE solver is $p$-th order accurate.

In other words, causal ODE distillation and causal CD share the same target; they differ only in how that target is approached---ODE distillation regresses to it via large jumps from $t$ to $0$ on pre-generated trajectories, while CD enforces it locally between adjacent timesteps on real data. This equivalence motivates causal CD as a principled substitute for causal ODE distillation.

\begin{figure}[t]
  \centering
  \hspace*{\figxshift}%
  \begin{subfigure}[t]{\leftfigwidth}
    \centering
    \includegraphics[width=\linewidth]{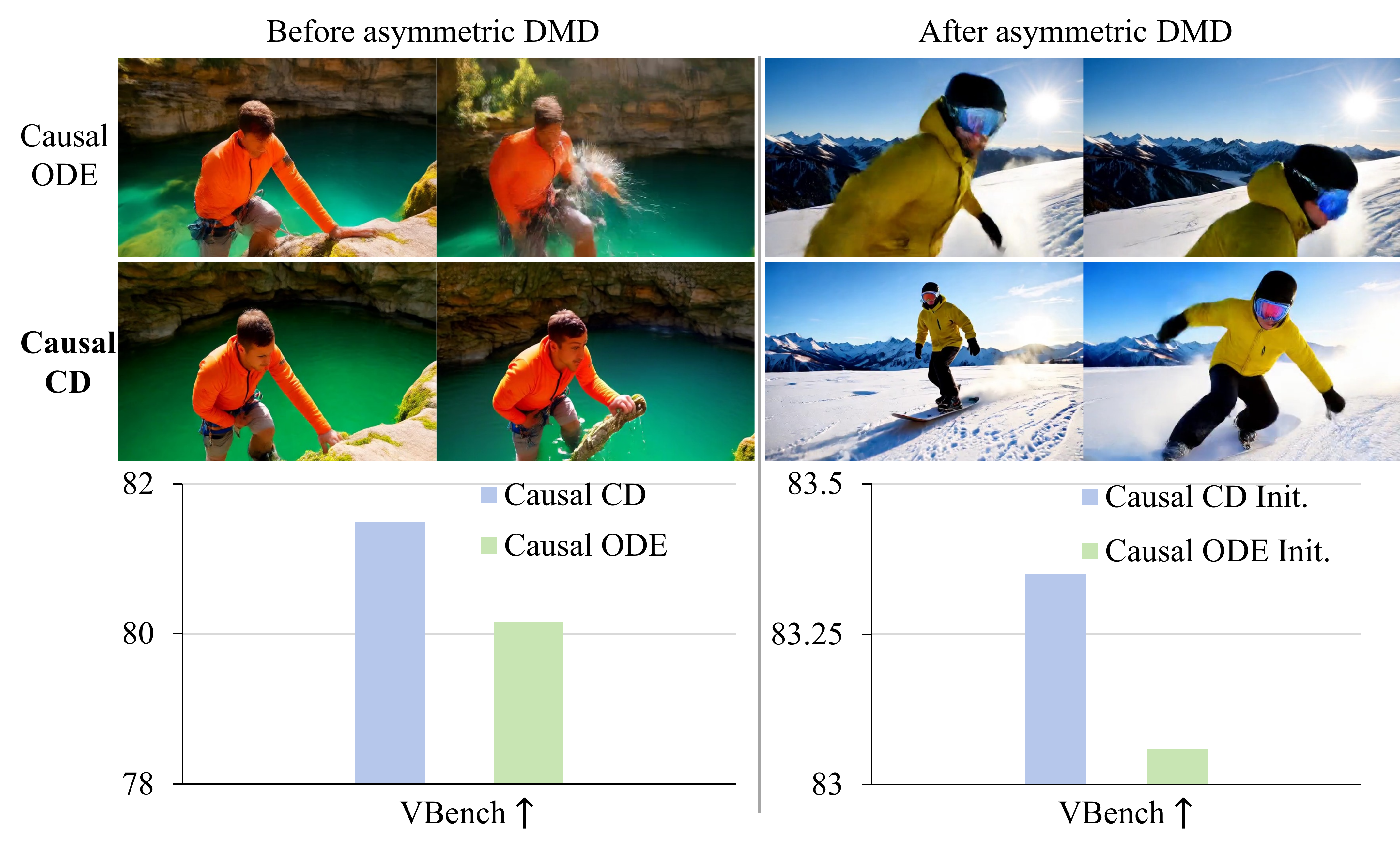}
    \caption{\footnotesize Performance comparison.
Causal CD achieves higher VBench scores than causal ODE both when directly evaluated at Stage~2 and when used as the initialization for Stage~3, while also producing videos with better visual quality. Init. denotes initialization.
}
    \label{fig:causal-cd-a}
  \end{subfigure}%
  \hspace{\figgap}%
  \begin{subfigure}[t]{\rightfigwidth}
    \centering
    \includegraphics[width=\linewidth]{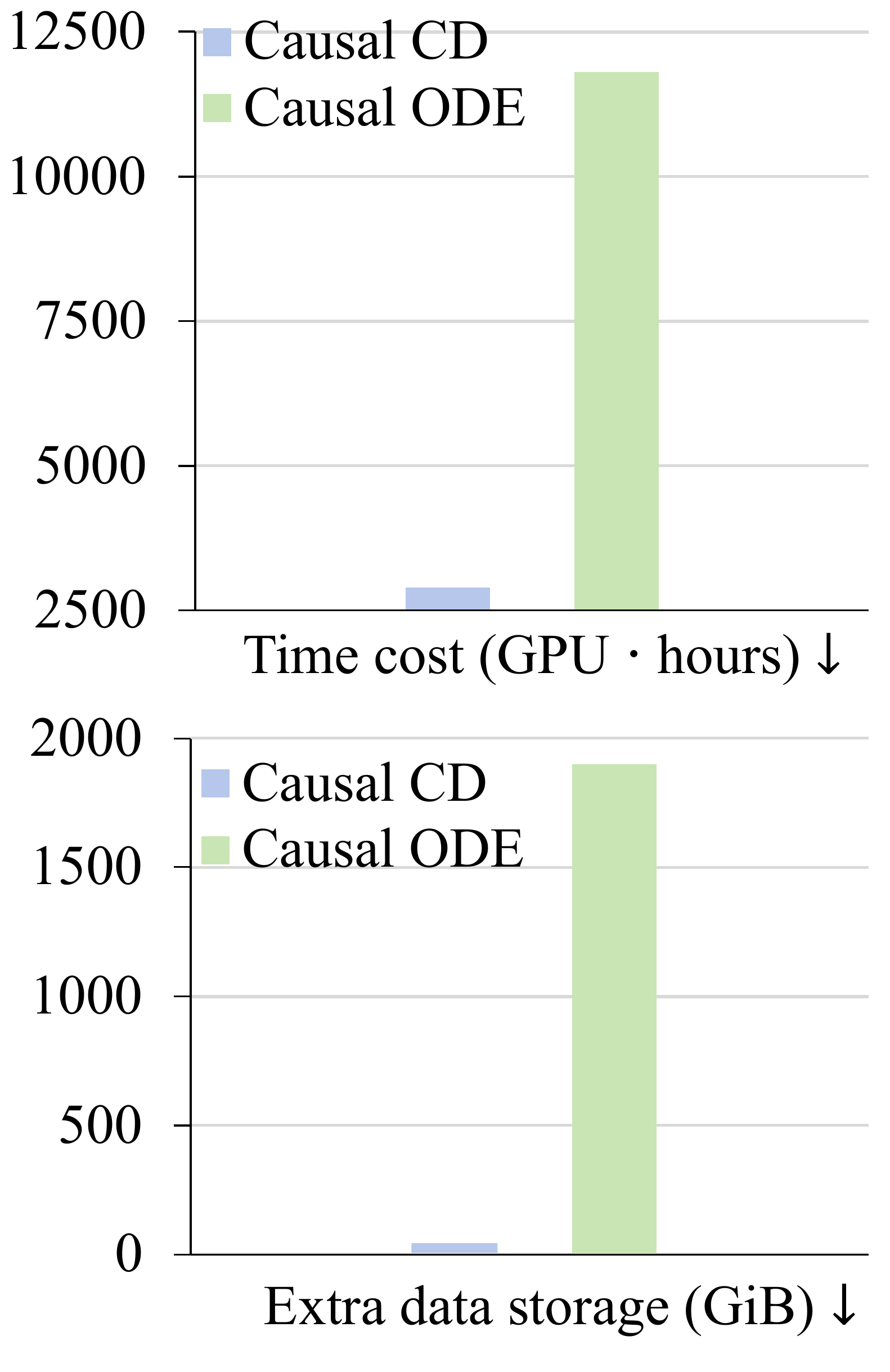}
    \caption{\footnotesize Training-efficiency.
Causal CD dramatically reduces training time and requires no extra storage.
}
    \label{fig:causal-cd-b}
  \end{subfigure}
\vspace{-.25cm}
  \caption{\textbf{Performance and efficiency comparison between causal CD and causal ODE.}
Causal CD outperforms causal ODE, while subsequently improving training efficiency in time and storage.}
  \label{fig:cd-is-better-than-ode}
\end{figure}

\paragraph{Causal CD is more efficient and yields a stronger initialization.} Beyond principled equivalence, Causal CD delivers two practical advantages over causal ODE distillation. \textbf{(1) Efficiency.} 
Causal ODE distillation requires the teacher to pre-generate a full multi-step PF-ODE trajectory for every training datum (e.g., 48 steps), store the paired trajectories offline, and regenerate them whenever the teacher or data distribution changes. Causal CD requires only \emph{one} teacher ODE step per training iteration, performed online on ground-truth videos. At our 80K-video scale (Fig.~\ref{fig:cd-is-better-than-ode}(b)), this reduces the data curation and training cost from $\sim$$11{,}600$ to $\sim$$2{,}900$ A800-GPU hours ($\sim$$4\times$ speedup) and the auxiliary storage from $\sim$$1{,}900$\,GiB to zero. \textbf{(2) Quality.} Causal ODE distillation regresses $\vx_t^i$ to $\vx_0^i$ across the entire $t \to 0$ interval, asking the student to bridge a large gap in one step. Causal CD instead pairs $\vx_t^i$ with $\hat{\vx}^i_{t-\Delta t}$ between adjacent timesteps, reducing the per-step gap to $\Delta t$ and yielding an easier optimization target---consistent with prior observations in bidirectional distillation~\cite{liu2023instaflow}. Empirically (Fig.~\ref{fig:cd-is-better-than-ode}(a)), causal CD matches or surpasses causal ODE distillation on VBench~\cite{huang2024vbench} both at the end of Stage~2 and after the asymmetric DMD stage. These advantages are structural properties of the local-pairing scheme and apply uniformly across chunk-wise and frame-wise configurations.

\paragraph{Causal Forcing++.} Putting these pieces together, our pipeline inherits Stage~1 (teacher forcing AR diffusion training) and Stage~3 (asymmetric DMD with self-rollout) from Causal Forcing~\cite{zhu2026causal}, and replaces its Stage~2 with the causal CD objective of Eq.~(\ref{eq:causal-cd}). We refer to this pipeline as \textbf{\textit{Causal Forcing++}}. In summary, Causal Forcing++ improves over prior AR diffusion distillation along two complementary axes. Relative to Self Forcing~\cite{huang2025self}, whose ODE initialization from a bidirectional teacher violates frame-level injectivity and degenerates to a conditional-expectation target, Causal Forcing++ is both \emph{correct}---targeting the correct AR flow map---and \emph{more efficient}, since causal CD avoids offline trajectory generation entirely. Relative to Causal Forcing~\cite{zhu2026causal}, whose causal ODE initialization recovers the correct target but is bottlenecked by offline trajectory generation, Causal Forcing++ is both \emph{more efficient} and \emph{higher-quality}, owing to the smaller per-step optimization gap of local consistency pairing.

\subsection{Application to Action-Conditioned World Models}
\label{sec:application}

We further showcase the applicability of our pipeline to action-conditioned world model generation. Following the Genie3-style~\cite{genie3} camera-pose conditioning paradigm, we instantiate a chunk-wise 4-step generator based on Wan2.1-1.3B, using camera poses as the action signal. The pipeline involves three stages: (i) constructing a camera-pose-annotated training dataset with WorldPlay~\cite{sun2025worldplay}; (ii) finetuning Wan2.1-1.3B into a bidirectional camera-pose-conditioned diffusion model by injecting pose information via PRoPE~\cite{li2026cameras}; and (iii) distilling this bidirectional model with Causal Forcing++ into an interactive action-conditioned world model. Qualitative results are shown in Fig.~\ref{fig:action}, and full demos are provided on the project page. Further reducing the action-conditioned variant to the frame-wise 2-step setting for fully real-time interaction is left for future work.

\newcommand{\ablationfigwidth}{\linewidth}

\begin{figure}
  \centering

  \begin{subfigure}{\ablationfigwidth}
    \centering
    \includegraphics[width=\linewidth]{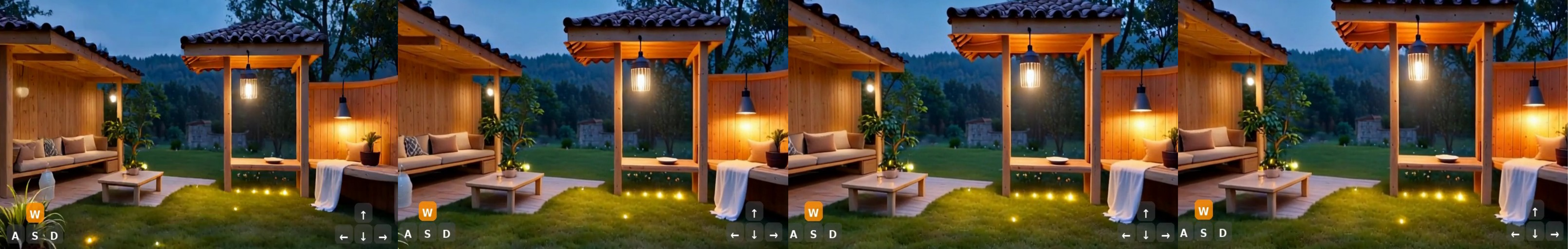}
    \caption{\footnotesize Action: moving forward continuously.}
  \end{subfigure}

  \vspace{0.5em}

  \begin{subfigure}{\ablationfigwidth}
    \centering
    \includegraphics[width=\linewidth]{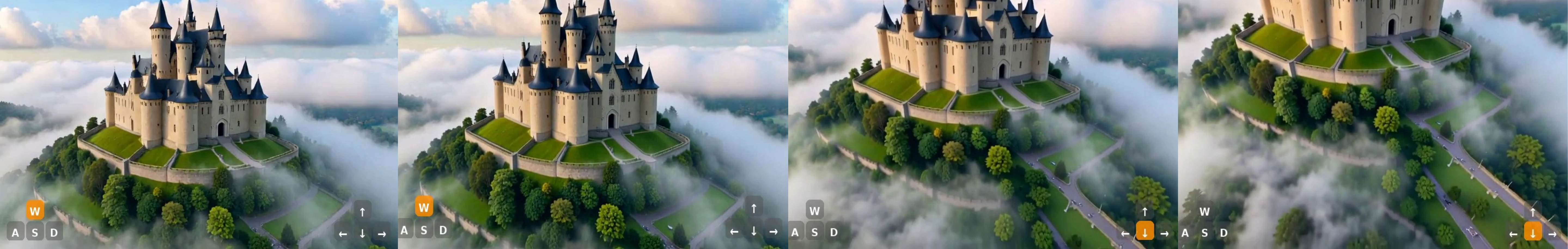}
    \caption{\footnotesize Action: move forward first, then tilt the camera downward.}
  \end{subfigure}

  \caption{\textbf{Application of Causal Forcing++ to the action-conditioned world model.} Causal Forcing++ enables efficient, high-quality distillation from a bidirectional model to a low-latency AR model, thereby enabling interactive generation toward a Genie3-style world model.}
  \label{fig:action}

\end{figure}

\subsection{Further Discussion}

\subsubsection{Whether Causal Score Distillation Yields Strong Initialization}

\label{sec: causal dmd}
Since causal consistency distillation has been shown to provide a strong few-step initialization above, a natural question is whether its counterpart in step distillation, namely score distillation~\cite{wang2023prolificdreamer,luo2023diff,yin2024one}, can also serve this role. Given that score distillation often yields higher quality than consistency distillation in bidirectional distillation~\cite{yin2024one,yin2024improved,zheng2025large}, this appears promising. In this section, we examine whether causal score distillation actually works.

\begin{figure}[t]
  \centering
  \hspace*{\figxshift}%
  \begin{subfigure}[t]{\leftfigwidth}
    \centering
    \includegraphics[width=0.95\linewidth]{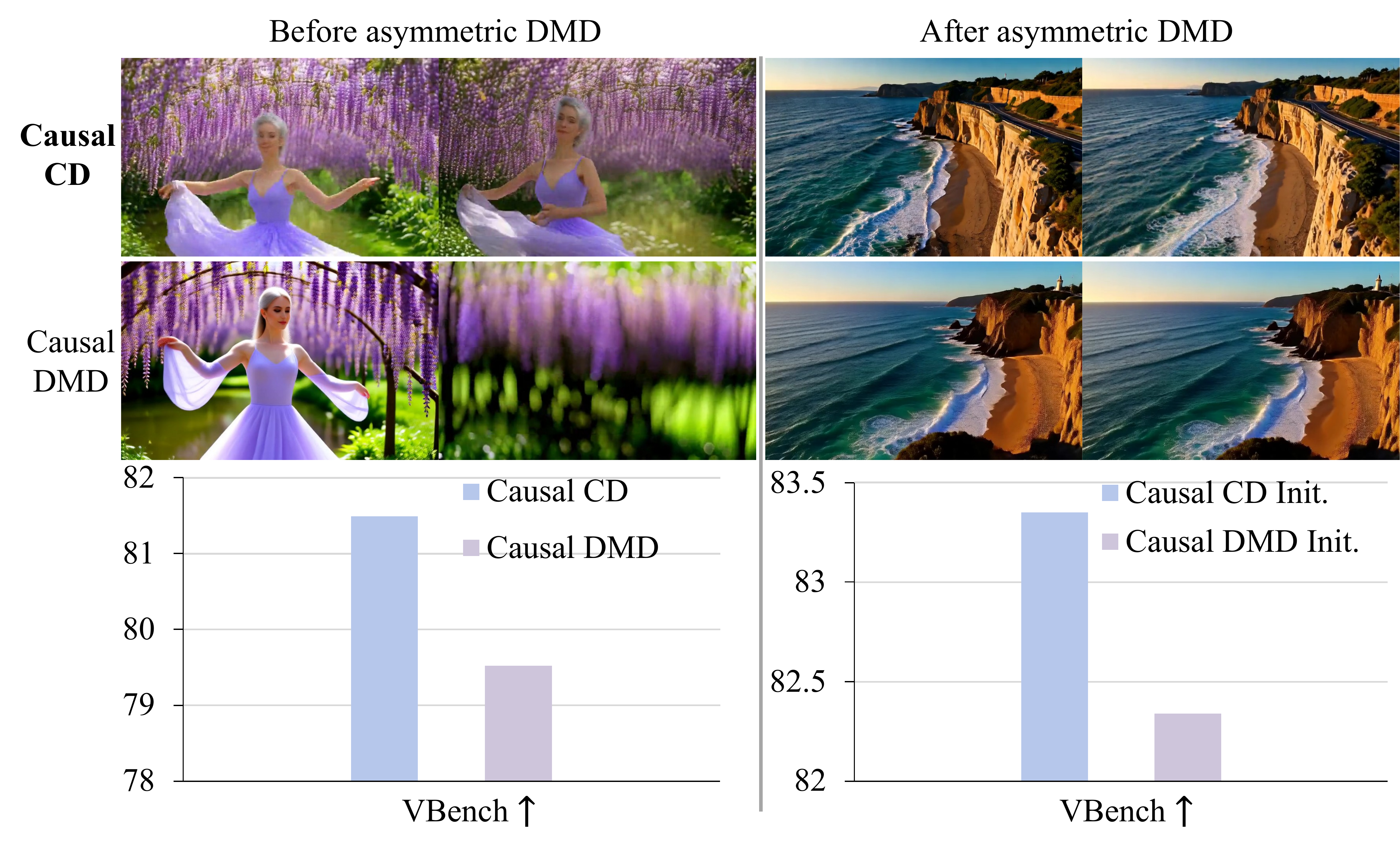}
    \caption{\footnotesize Performance comparison. Causal DMD yields lower VBench scores than causal CD both when directly evaluated after Stage 2 and when used as the initialization for Stage 3. Although causal DMD produces sharper early frames, its quality rapidly drifts during autoregressive rollout, leading to severe exposure bias. Init. denotes initialization.}
    \label{fig:causal-dmd-a}
  \end{subfigure}%
  \hspace{\figgap}%
  \begin{subfigure}[t]{\rightfigwidth}
    \centering
    \includegraphics[width=1.25\linewidth]{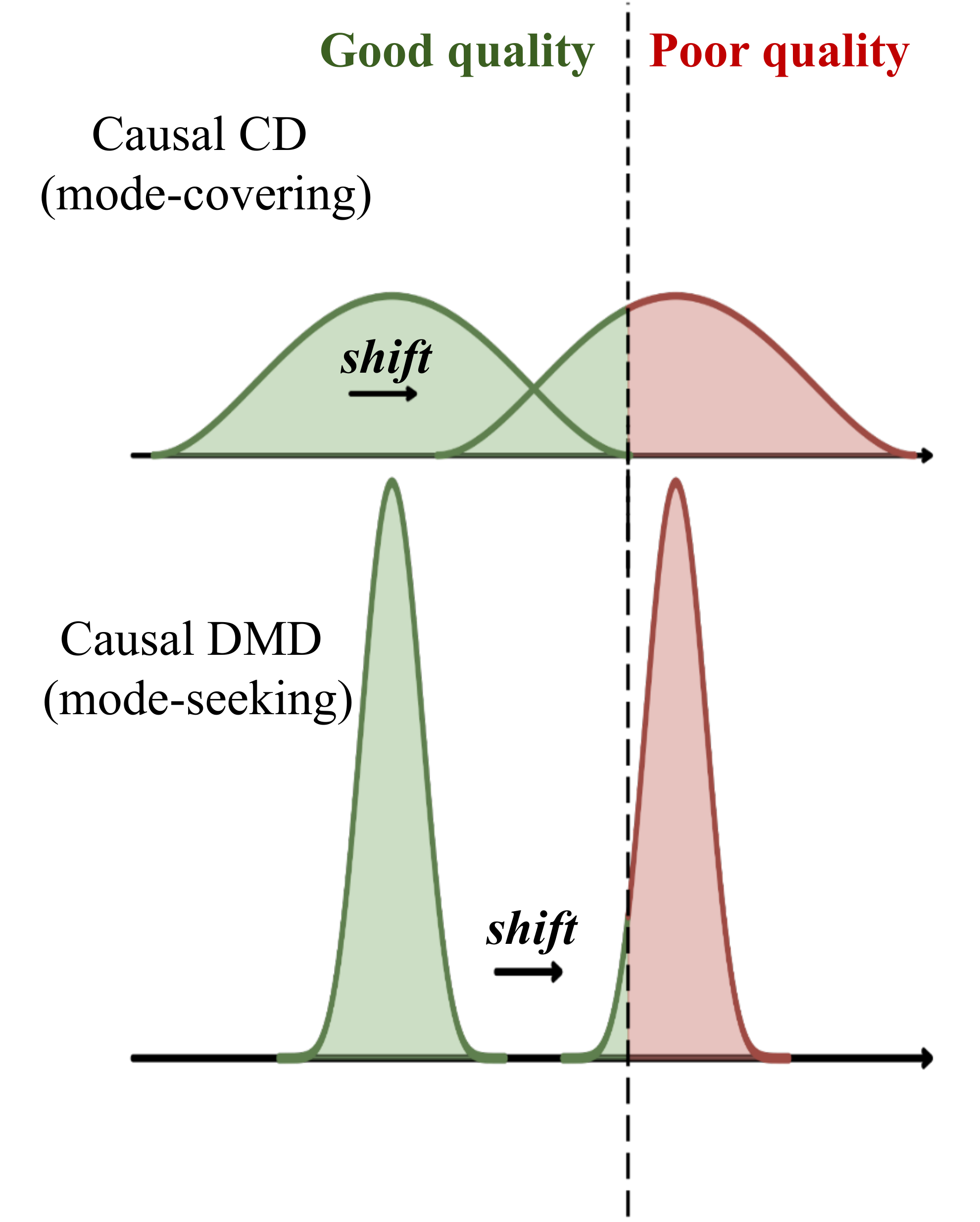}
    \caption{\footnotesize Under history shift, causal CD preserves good-quality mass; causal DMD shifts most mass into poor-quality regions.}
    \label{fig:causal-dmd-b}
  \end{subfigure}

  \caption{\textbf{Performance comparison between causal CD and causal DMD and the intuitive explanation.} Causal DMD achieves better early-frame quality than causal CD, but suffers from severe exposure bias in later frames, since mode-seeking DMD is more sensitive to accumulated errors.}
  \label{fig:dmd-is-worse-than-cd}
\end{figure}

\paragraph{Causal DMD with teacher forcing.} We begin by adapting score distillation into a teacher forcing causal form. Specifically, we take DMD~\cite{wang2023prolificdreamer,luo2023diff,yin2024one,yin2024improved} as a representative example and examine its causal variant. DMD optimizes the student generator through the gradient of the KL divergence between the noised student and data distributions. The student first produces $\tilde{\vx}\sim p_\theta(\tilde{\vx})$, which is then perturbed to $\tilde{\vx}_t$ through the forward diffusion process, inducing $p_{\theta,t}(\tilde{\vx}_t)$. A frozen diffusion model $s_{\text{real}}$ estimates the data score, while an online-trained diffusion model $s_{\text{fake}}$ estimates the student score. Their difference provides the distribution-matching direction, which is backpropagated to update the generator as follows:
\begin{align}
\label{eq:dmd}
\nabla_\theta\E_t[\KL(p_{\theta,t}(\tilde{\vx}_t)||p_{\text{data},t}(\tilde{\vx}_t))]
   = -\E_{\tilde{\vx}, t, \tilde{\vx}_t}[(s_\text{real}(\tilde{\vx}_t,t)-s_\text{fake}(\tilde{\vx}_t,t))\frac{\partial \tilde{\vx}}{\partial \theta}].
\end{align}
When distilling an AR diffusion model, the setting becomes slightly different: the teacher models $s_\text{real}$, $s_\text{fake}$ and the student $G_\theta$ all model conditional distributions autoregressively. Therefore, the above objective must be trained under teacher forcing, as follows:
\begin{align}
\label{eq:causal-dmd}
\nabla_\theta\E_t[\KL(p_{\theta,t}(\tilde{\vx}_t^i|\vx_{\mathrm{gt}}^{<i})||p_{\text{data},t}(\tilde{\vx}_t^i|\vx_{\mathrm{gt}}^{<i}))]
   = -\E_{\vx_{\mathrm{gt}}^{<i},\tilde{\vx}^i, t, \tilde{\vx}_t^i}[(s_\text{real}(\tilde{\vx}_t^i,\vx_{\mathrm{gt}}^{<i},t)-s_\text{fake}(\tilde{\vx}_t^i,\vx_{\mathrm{gt}}^{<i},t))\frac{\partial \tilde{\vx}^i}{\partial \theta}].
\end{align}
With this training scheme, we can directly distill an AR diffusion model via DMD into an AR few-step generator. Note that all models involved in causal DMD, including the teachers and the student, are autoregressive. It is therefore a purely causal distillation procedure, fundamentally different from the subsequent asymmetric DMD, where the teacher is bidirectional while the student is autoregressive. In summary, these two DMD stages serve distinct roles: causal DMD provides a few-step AR student initialization for the subsequent asymmetric DMD.

\paragraph{Severe exposure bias undermines causal DMD.}
Although bidirectional DMD usually achieves better quality than bidirectional CD~\cite{yin2024one,yin2024improved,zheng2025large}, we observe the opposite trend in the AR setting: causal DMD yields lower overall quality than causal CD, as shown by the Stage~2 VBench comparison in Fig.~\ref{fig:dmd-is-worse-than-cd}(a). More interestingly, this degradation is not uniform across frames. In the Stage~2 visualizations of Fig.~\ref{fig:dmd-is-worse-than-cd}(a), the first few frames generated by causal DMD even appear sharper and better than those from causal CD. However, as autoregressive generation proceeds, the later frames from causal DMD rapidly drift, accompanied by severe camera shifts, and eventually degrade to an unacceptable level. In contrast, although causal CD also suffers from gradual quality decay during autoregressive rollout, its degradation is far less severe. The Stage~3 results in Fig.~\ref{fig:dmd-is-worse-than-cd}(a) further show that using causal DMD as the initialization for the subsequent asymmetric DMD performs worse than using causal CD initialization. In summary, causal DMD suffers from substantially stronger exposure bias and is therefore unsuitable as a few-step initialization.

We further analyze why causal DMD can outperform CD in the first few frames but then rapidly suffers from amplified exposure bias, with an intuitive explanation illustrated in Fig.~\ref{fig:dmd-is-worse-than-cd}(b). As noted in prior work~\cite{zheng2025large}, DMD optimizes a reverse KL objective, whereas CD optimizes a forward KL objective. Compared with the mode-covering behavior induced by forward KL, the mode-seeking behavior of reverse KL tends to concentrate probability mass around high-density modes, leading to a sharper and less diverse distribution. This contrast is illustrated by the broader causal-CD distributions and the sharper causal-DMD distributions in Fig.~\ref{fig:dmd-is-worse-than-cd}(b). Early in autoregressive rollout, this concentration reduces the probability assigned to low-quality regions and thus improves generation quality, explaining the strong quality of the first few frames. However, such a concentrated distribution is also more sensitive to accumulated errors in autoregressive conditional generation.

Specifically, the arrows in Fig.~\ref{fig:dmd-is-worse-than-cd} denote the shift of the conditional distribution caused by history drift, i.e., the deviation of the self-generated prefix from the ground-truth history. Under such drift, the conditional distributions induced by both mode-covering causal CD and mode-seeking causal DMD shift toward the poor-quality region. For mode-seeking DMD, since the probability mass is highly concentrated, once the shifted mode moves into the poor-quality region, most generated samples tend to follow this erroneous mode, as shown by the shifted red DMD peak in Fig.~\ref{fig:dmd-is-worse-than-cd}(b). In contrast, mode-covering CD maintains a more dispersed distribution; even after the shift, a substantial portion of its probability mass can remain in the good-quality region, as shown by the broader shifted CD distribution in Fig.~\ref{fig:dmd-is-worse-than-cd}(b). Therefore, intuitively, mode-seeking DMD is more sensitive to accumulated history errors, causing exposure bias to rapidly amplify in later frames, whereas mode-covering CD suffers from exposure bias with much lower sensitivity.

\subsubsection{Distinction from Other AR Distillation Paradigms}
APT2~\cite{lin2026autoregressive} represents another line of work that is substantially different from the classical asymmetric DMD-based AR diffusion distillation paradigm. When initializing its GAN~\cite{goodfellow2014generative} generator, APT2 adopts teacher-forcing CD, which is highly related to our causal CD initialization. 

However, the two works significantly differ in both motivation and technical design. First, Causal Forcing++ is motivated by reducing the cost of causal ODE initialization in Causal Forcing while preserving theoretical equivalence, whereas Causal Forcing is the first work to theoretically show why such causal initialization is needed. In contrast, APT2 adopts teacher-forcing CD heuristically, without a corresponding theoretical motivation. Second, Causal Forcing++ remains within the asymmetric DMD framework of CausVid and Self Forcing: it uses a bidirectional diffusion model as the DMD teacher and supports temporal KV-cache inference. APT2, by contrast, is built as a GAN-based method, has no bidirectional-teacher formulation, does not use temporal KV cache, and instead concatenates noise with conditions. Built upon these distinct training algorithms and model architectures, this work open-sources the first asymmetric-DMD-driven few-step AR model initialized by causal CD. Since APT2 is not open-source, we do not compare with it below.
\section{Experiments}
\subsection{Setup}
\label{sec: setup}

\paragraph{Implementation details.} Following Causal Forcing~\cite{zhu2026causal}, we adopt Wan2.1-1.3B~\cite{wan2025wan} as the base model from which the AR few-step model is derived. The model generates videos at a resolution of 480 $\times$ 832 with 81 frames, using frame-wise autoregressive generation. We follow the three-stage distillation pipeline of Causal Forcing, namely: Stage 1, teacher forcing AR diffusion training; Stage 2, few-step initialization; and Stage 3, asymmetric DMD with self rollout. For Stage 2 causal CD, we use the square norm, adopt 48 discretized timesteps, and employ the Euler solver.
 For the number of sampling steps, in Stage 3 we train 4-step ($t=1,0.9375,0.8333,0.625$), 2-step ($t=1,0.8333$), and 1-step settings. For settings with fewer than 4 steps, we follow ASD~\cite{yang2025towards} and keep the generation of the first latent frame at 4 steps, while using 2 steps or 1 step, respectively, for the subsequent 20 latent frames. Note that this trick is used only in Stage 3 and is not involved in the first two stages. All teacher and student models used throughout the pipeline are based on Wan2.1-1.3B, except for the real score model in the Stage 3 DMD, which is based on Wan2.1-14B. For Stage 1,2, we utilize an 80K dataset including videos sampled from OpenVid~\cite{nan2024openvid}; for Stage 3, we use the  VidProM~\cite{wang2024vidprom} dataset. The numbers of training steps for the three stages are 20K, 5K, and 1K, respectively, and the batch size is 64. All other hyperparameters are kept the same as in Causal Forcing. 

\paragraph{Evaluation.} Following Causal Forcing, we adopt two benchmarks: VBench~\cite{huang2024vbench} and VisionReward~\cite{xu2024visionreward}. For VBench, we report the overall metrics and additionally evaluate dynamic degree separately using the 100 prompts from Causal Forcing. For VisionReward, we also use the 100 prompts from Causal Forcing and report both the overall score and instruction-following performance. For readability, all metrics are multiplied by 100. All other evaluation details follow Causal Forcing. In addition, we report first-frame latency and throughput. \textit{\textbf{These efficiency metrics are measured \emph{on the single A800 GPU without the VAE-related time cost}, rather than on H100 as in the Self Forcing and Causal Forcing papers.}}

\subsection{Results}
\begin{table}[h]
\small\setlength{\tabcolsep}{1pt} 
\caption{\textbf{Comparisons with existing methods.} Our method reduces latency by half and improves throughput, while achieving generation quality comparable to or even better than previous SOTA methods. Dynamic., Vision. and Instruct. denote Dynamic Degree, VisionReward and Instruction Following, respectively. Throughput and latency are measured in FPS and seconds on an \textbf{\textit{A800}}.\protect\footnotemark}
  \label{tab:performance-comparison}
  \centering
  \begin{tabular}{lcccccccc}
      \toprule
      Model & Throughput $\uparrow$& Latency $\downarrow$&Total$\uparrow$ & Quality$\uparrow$ & Semantic$\uparrow$ 
      & Dynamic.$\uparrow$
      & Vision.$\uparrow$
      & Instruct.$\uparrow$
       \\
    \midrule
    CausVid~\cite{yin2025slow}& 10.4 & 0.60 & 81.33 & 83.98 & 70.72 &  62  & 5.741 & 12 \\
    Self Forcing~\cite{huang2025self} & 10.4 & 0.60 & 83.74 & 84.48 & 80.77 & 57 & 5.820 & 48 \\
   Causal Forcing~\cite{zhu2026causal}  & 10.4 & 0.60 & 84.04 & 84.59& \textbf{81.84} & \underline{\textbf{\textit{68}}} & 6.326 & \textbf{56} \\
      \textbf{Causal Forcing++} (1-step)  & \textbf{20.7} & \textbf{0.27} &83.35 & 84.50  &   78.75 & 66 & 5.412 & 38\\
      \textbf{Causal Forcing++} (2-step)  & \underline{\textit{\textbf{14.1}}} & \textbf{0.27} & \textbf{84.14} & \underline{\textbf{\textit{84.89}}}  &   \underline{\textbf{\textit{81.13}}} & 64 &\underline{\textbf{\textit{6.661}}} & \underline{\textbf{\textit{51}}} \\
     \textbf{Causal Forcing++} (4-step)  & 8.69 & \textbf{0.27} & \underline{\textbf{\textit{84.10}}} & \textbf{84.94}  &   80.75 & \textbf{71} &\textbf{6.798} &  47\\
    \bottomrule
  \end{tabular}\\
\end{table}
\footnotetext{Because we adopt the ASD trick~\cite{yang2025towards}, the first-frame latency for 1-step, 2-step, and 4-step generation is identical, and is determined by the 4-step generation of the first frame.}
\begin{figure}[htbp]
    \centering
    \includegraphics[width=\textwidth]{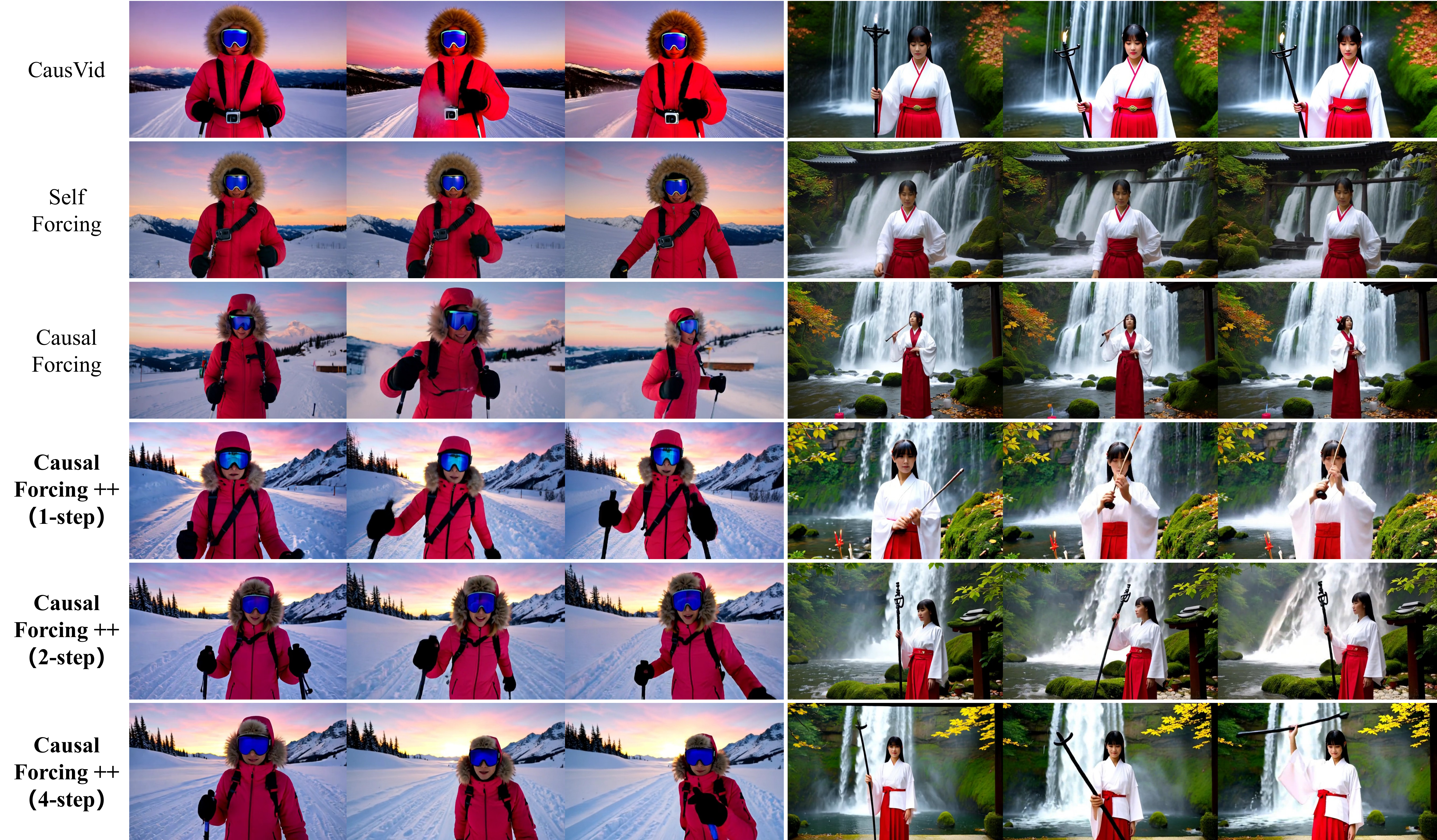}
    \caption{\textbf{Performance comparison.} Our Causal Forcing++ achieves quality and dynamics comparable to or even surpassing Causal Forcing, while outperforming CausVid and Self Forcing.}
    \label{fig:performance-comparison}
\end{figure}

\paragraph{Performance comparison with existing methods.} We compare Causal Forcing++ with prior AR diffusion distillation methods, including CausVid~\cite{yin2025slow}, Self Forcing~\cite{huang2025self}, and Causal Forcing~\cite{zhu2026causal}, which are all chunk-wise models with 4-step sampling. 

As illustrated in Tab. \ref{tab:performance-comparison}, our frame-wise Causal Forcing++ (2-step generation) achieves the best VBench total score. Its Total score (84.14) and Quality score (84.89) surpass all previous methods, while its Semantic score remains comparable to Causal Forcing. Notably, while maintaining overall generation quality comparable to, and in some aspects better than, prior state-of-the-art (SOTA) methods, our 2-step frame-wise Causal Forcing++ improves throughput by about 1.4× and reduces latency by 50\%. Furthermore, Causal Forcing++ under 4-step generation attains even higher Quality and VisionReward scores, and achieves a Dynamic Degree beyond all previous SOTA methods, while still maintaining 50\% lower first-frame latency.

These quantitative results are consistent with the visualization in Fig. \ref{fig:performance-comparison}. CausVid and Self Forcing both exhibit weaker dynamics; moreover, CausVid shows over-saturation artifacts, while Self Forcing suffers from object inconsistency, as illustrated by the spurious horizontal structure growing from the building next to the waterfall in the right-hand demo. In contrast, under fewer-step frame-wise generation, Causal Forcing++ attains quality and dynamics comparable to those of Causal Forcing, and even shows advantages in aesthetic aspects such as color vividness and brightness.

\begin{figure}
  \centering

  \begin{subfigure}{\ablationfigwidth}
    \centering
    \includegraphics[width=\linewidth]{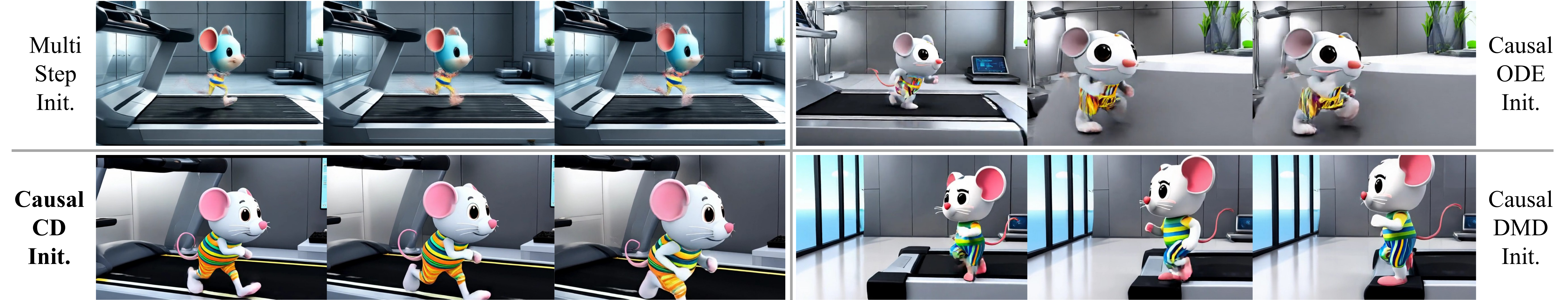}
    \caption{\footnotesize \textbf{Final results of 1-step asymmetric DMD under different initializations.} AR diffusion initialization results in severe blur and almost no motion, causal ODE suffers from scene collapse, and causal DMD blurs the mouse’s legs into a single indistinguishable mass. In contrast, causal CD initialization preserves both strong dynamics and high visual quality.  Init. denotes initialization.}
  \end{subfigure}

  \vspace{0.5em}

  \begin{subfigure}{\ablationfigwidth}
    \centering
    \includegraphics[width=\linewidth]{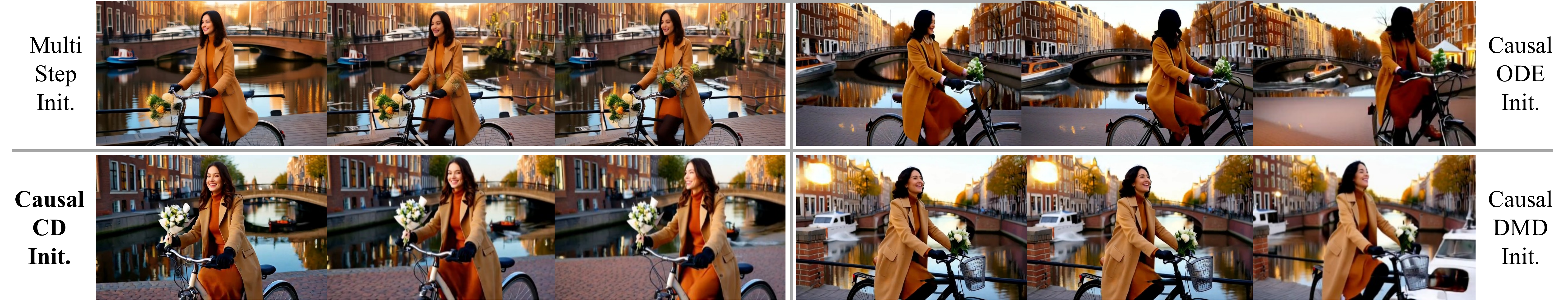}
    \caption{\footnotesize \textbf{Final results of 2-step asymmetric DMD under different initializations.} Multi-step init. blurs the bouquet, causal ODE init. turns the person’s head black, and causal DMD init. introduces a bright spot in the first frame and a boat in later frames. In contrast, causal CD init. maintains stable quality. Init. denotes initialization.}
  \end{subfigure}

  \vspace{0.5em}

  \begin{subfigure}{\ablationfigwidth}
    \centering
    \includegraphics[width=\linewidth]{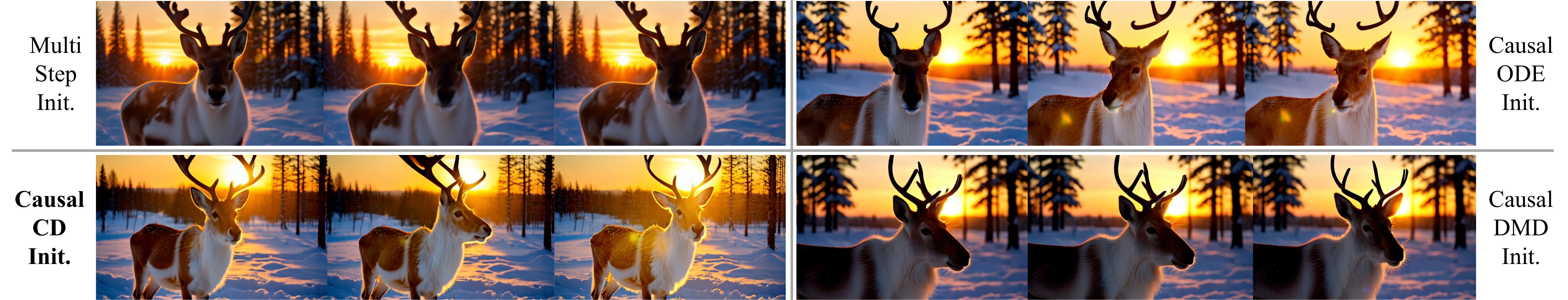}
    \caption{\footnotesize \textbf{Final results of 4-step asymmetric DMD under different initializations.} Both multi-step and causal DMD initializations produce poor visual quality. Causal ODE exhibits antler separation artifacts, whereas causal CD initialization yields high-quality results. Init. denotes initialization.}
  \end{subfigure}

  \caption{\textbf{Visual comparisons after asymmetric DMD with different initializations.} Causal CD achieves results comparable to or even better than causal ODE, whereas multi-step and causal DMD initializations perform worse.}
  \label{fig:ablation}

\end{figure}

\paragraph{Ablation studies.} 
In this section, we provide a detailed method ablation. We experiment under the frame-wise 1, 2, and 4-step settings, respectively, and evaluate Self Forcing~\cite{huang2025self} ODE initialization, multi-step AR diffusion initialization, causal ODE initialization, causal DMD initialization, and our causal CD initialization. Note that the difference in the number of generation steps only applies to Stage 3, while Stage 2 remains identical across settings, with special note that the causal ODE is trained under the 4-step setting. In addition to the quality metrics in Sec.~\ref{sec: setup}, we also list here the training time (measured in A800 GPU$\cdot$ hours,  rounded to the nearest hundred) for Stage 2 (with ODE including the 80K data curation time) and the extra storage for the ODE paired data (measured in GiB). The quantitative results and visual comparisons are illustrated in Tab. \ref{tab:ablation} and Fig. \ref{fig:ablation}, respectively, from which we draw the following conclusions.

\paragraph{Self Forcing ODE initialization performs consistently poorly in the frame-wise setting.} Specifically, under the 2-step and 4-step settings, Self Forcing ODE initialization underperforms all other methods across all metrics. Its VBench Total score remains below 80, and its Dynamic Degree reaches at most 2. Beyond its poor quality, the ODE stage also costs 5,000 A800 GPU$\cdot$ hours and requires 1,500 GB of storage.

\paragraph{A strong few-step initialization is indispensable, especially in aggressive low-step settings.}
Using the multi-step AR diffusion model directly as initialization yields consistently weak performance. This issue is most severe in the 1-step setting, where the model nearly collapses in dynamics and instruction following, achieving only 0 Dynamic Degree, 1.101 VisionReward, and even a -14 Instruction Following score. Although the performance improves as the number of steps increases, it still remains clearly inferior even under 4-step generation, where its VisionReward is only 5.166, still far below all explicit few-step initialization methods. This shows that AR diffusion initialization is insufficient, and that explicit few-step adaptation before asymmetric DMD is crucial, corroborating Sec. \ref{sec:necessity}.

\paragraph{Causal CD is a stronger and more efficient substitute for causal ODE.}
Causal CD matches or surpasses causal ODE across all step settings while dramatically reducing Stage 2 cost. In 1-step generation, it improves Total, Quality, and Dynamic Degree over causal ODE, with comparable VisionReward and Instruction Following. In the 2-step setting, it achieves the best overall performance, with the highest Total, Quality, and VisionReward scores, while maintaining a comparable Semantic score. In the 4-step setting, causal CD again gives the strongest overall results. Meanwhile, causal ODE requires around 11,600 A800 GPU$\cdot$ hours and 1,900\,GiB extra storage for ODE-paired data curation and training, whereas causal CD requires only around 2,900 A800 GPU$\cdot$ hours and no extra storage. Thus, causal CD reduces Stage 2 time cost by roughly 4$\times$, completely removes storage overhead, and serves as a better substitute for causal ODE initialization, consistent with Sec.~\ref{sec:substitute}.

\paragraph{Causal DMD is not an ideal replacement for few-step initialization.} Although causal DMD improves substantially over AR diffusion initialization, it is weaker than causal CD and is usually also worse than causal ODE. For example, its VisionReward remains about 0.5 lower than that of causal CD initialization across all settings. This suggests that causal DMD initialization is suboptimal.

These ablation results are highly consistent with Sec. \ref{sec: method}, and provide strong evidence for the validity and superiority of replacing causal ODE with causal CD in our Causal Forcing++.

\begin{table}
\small\setlength{\tabcolsep}{1pt} 
  \caption{\textbf{Ablation study.} Causal CD is comparable to or even outperforms causal ODE, while dramatically reducing both time and storage cost. In contrast, Self Forcing ODE initialization, multi-step and causal DMD initialization perform worse.}
  \label{tab:ablation}
  \centering
  \begin{tabular}{lcccccccc}
      \toprule
      Model & Total$\uparrow$ & Quality$\uparrow$ & Semantic$\uparrow$ 
      & Dynamic.$\uparrow$
      & Vision.$\uparrow$
      & Instruct.$\uparrow$
      & Time (Stage 2)$\downarrow$
      & Extra Storage $\downarrow$
       \\
    \midrule
     \rowcolor{catgray}
    \multicolumn{9}{l}{\textit{1-Step Asymmetric DMD}}\\
        Self Forcing initialization  & 78.87 & 79.85  &  74.95 & 0 & 1.992 & -12 &\cellcolor{red!10}5000 &\cellcolor{red!10}1500\\
      AR diffusion initialization  & 80.54 & 80.97  &   78.84& 0 & 1.101 & -14 &\cellcolor{green!10}- &\cellcolor{green!10}0\\
      Causal ODE initialization  &\underline{\textbf{\textit{83.06}}} & \underline{\textbf{\textit{83.88}}}  &   \textbf{79.77} & 46 & \textbf{5.464} & \textbf{40} & \cellcolor{red!10}11600 & \cellcolor{red!10}1900 \\
      Causal DMD initialization  &82.34 & 83.50  &   77.71 & \underline{\textbf{\textit{62}}} & 4.868 & 20&\cellcolor{green!10}2900 &\cellcolor{green!10}0\\
      Causal CD initialization  &\textbf{83.35} &\textbf{ 84.50}  &   \underline{\textbf{\textit{78.75}}} & \textbf{66} & \underline{\textbf{\textit{5.412}}} & \underline{\textbf{\textit{38}}} & \cellcolor{green!10}2900 &\cellcolor{green!10}0\\
        \midrule
     \rowcolor{catgray}
    \multicolumn{9}{l}{\textit{2-Step Asymmetric DMD}}\\
     Self Forcing initialization  & 79.44 & 80.43  &   75.47 & 0 & 2.826 & -14 &\cellcolor{red!10}5000 &\cellcolor{red!10}1500\\
      AR diffusion initialization  & 82.43 & 83.04  &   80.00& 8 & 3.645 & 16 & \cellcolor{green!10}-& \cellcolor{green!10}0 \\
      Causal ODE initialization  &\underline{\textbf{\textit{83.77}}} & 84.42  &   \textbf{81.19}& 57 &\underline{\textbf{\textit{6.224}}} & \underline{\textbf{\textit{46}}}&\cellcolor{red!10}11600& \cellcolor{red!10}1900 \\
      Causal DMD initialization  & 83.73 & \underline{\textbf{\textit{84.56}}}  &   80.39& \textbf{69} & 6.108 & 38 & \cellcolor{green!10}2900&\cellcolor{green!10}0 \\
      Causal CD initialization  &\textbf{84.14} & \textbf{84.89}  &   \underline{\textbf{\textit{81.13}}} & \underline{\textbf{\textit{64}}} &\textbf{6.661} & \textbf{51} & \cellcolor{green!10}2900& \cellcolor{green!10}0 \\
        \midrule
     \rowcolor{catgray}
    \multicolumn{9}{l}{\textit{4-Step Asymmetric DMD}}\\
     Self Forcing initialization  & 79.82 & 80.58  &  76.76& 2 & 3.675 & -2 &\cellcolor{red!10}5000 &\cellcolor{red!10}1500\\
      AR diffusion initialization  & 83.41 & 84.42  &   \underline{\textbf{\textit{79.38}}} & 62 & 5.166 &  30& \cellcolor{green!10}- &\cellcolor{green!10}0\\
      Causal ODE initialization  & \underline{\textbf{\textit{83.78}}} & \underline{\textbf{\textit{84.90}}}  &   79.28  &  \textbf{75}&\underline{\textbf{\textit{6.435}}} &  \underline{\textbf{\textit{42}}}& \cellcolor{red!10}11600 & \cellcolor{red!10}1900 \\
      Causal DMD initialization  &83.49 & 84.84  &   78.11& 67 & 6.309&  36 & \cellcolor{green!10}2900 & \cellcolor{green!10}0\\
      Causal CD initialization  & \textbf{84.10} & \textbf{84.94}  &   \textbf{80.75} & \underline{\textbf{\textit{71}}} &\textbf{6.798} &  \textbf{47}& \cellcolor{green!10}2900 & \cellcolor{green!10}0\\
    \bottomrule
  \end{tabular}
\end{table}

\section{Conclusion}

In this paper, we identify the training inefficiency of existing AR diffusion distillation methods and propose Causal Forcing++, which replaces causal ODE distillation with causal consistency distillation. Our method substantially improves training efficiency and, for the first time, achieves performance comparable to or better than prior SOTA methods under the frame-wise 2-step setting, reducing latency by 50\% . Experiments show that our method achieves strong performance, demonstrating its effectiveness.

\clearpage

{
  \small
  \bibliographystyle{unsrt}
  \bibliography{main}
}

\end{document}